# Balanced Virtual Humans Interacting with their Environment

Antoine Rennuit[123], Alain Micaelli[1], Xavier Merlhiot[1], Claude Andriot[1], François Guillaume[2], Nicolas Chevassus[2], Damien Chablat[3], Patrick Chedmail[3]
[1]CEA\LIST, Fontenay-aux-Roses, France
[2]EADS\CCR, Suresnes, France
[3]IRCCyN, Nantes, France

## 1. Introduction

The animation of human avatars seems very successful; the computer graphics industry shows outstanding results in films everyday, the game industry achieves exploits... Nevertheless, the animation and control processes of such manikins are very painful. It takes days to a specialist to build such animated sequences, and it is not adaptive to any type of modifications.

Our main purpose is the virtual human for engineering, especially virtual prototyping. As for this domain of activity, such amounts of time are prohibitive.

We focus our work on interactive virtual human enabling to drive avatars in real time thanks to motion capture devices. Unfortunately, at the moment the quality of the animations produced is far from the quality obtained by artists in the computer graphics area. We aim at filling the gap.

Thus we proposed the architecture given Fig. 1, which distinguishes, and splits the simulation from the control itself. This architecture, which we introduced in [1], is innovative for the virtual reality domain, and was borrowed from robotics.

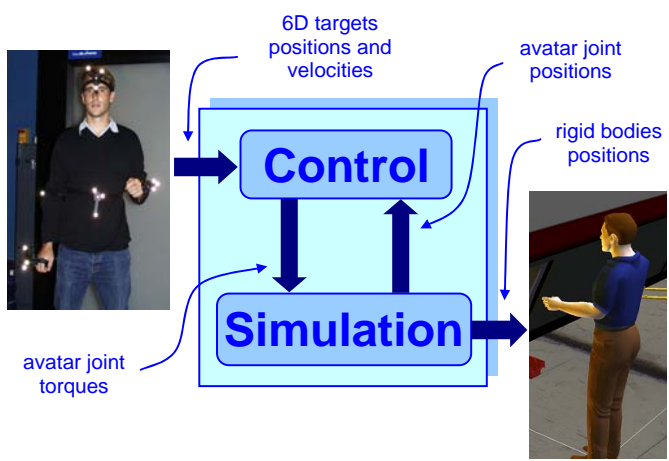

**Fig. 1 Global scheme of our system**

Many problems must be solved to succeed in. All the problems arise because of the difference between the real world, and the simulated world. This divergence takes two main shapes: the difference in terms of the human, and the difference in terms of the environment of both worlds.

**Difference of morphology:**
*Not conflicting*

The first variation can be easily illustrated, for example when the actor, and its avatar are differently morphologied. Let's take a giant actor, trying to control a dwarf virtual manikin. We want the hands and the feet of the avatar to track the hands and feet of its real world counterpart. As the actor is much taller than its avatar, there are situations when the avatar cannot reach all the targets at the same time; this kind of problems is called *conflicting retargeting*.

The retargeting problem was firstly addressed by Gleicher in [2]. His work makes it able to retarget the movement of a character onto another differently morphologied character. His own approach is purely kinematical, but another method by Popovic and Witkin [3], regard dynamics equations as a constraint, and thus preserve the physical nature of the movement. Unfortunately, because of an optimization that is made on the whole motion at once (this technique is called space-time optimization), the method is unuseful to our purpose: the entire movement must be known before retargeting. Moreover these techniques solve the retargeting problem only when the retargeting is not conflicting…

*Conflicting*

Another family of methods solving the retargeting problem was introduced by Baerlocher and Boulic in [4]. They noticed that in case of conflicting situations some targets are more important than others, e.g. one always want to keep feet on the ground (otherwise the manikin would seem flying), whereas hands are a bit less important.

This problem of conflicting retargeting, can express itself in trickier ways than the only impossibility to reach all targets.

Let's take again our example of the giant and the dwarf. Before the moment when all targets are not reachable at same time anymore, the virtual dwarf enters a state where kinematically (or geometrically) speaking it can reach his targets, but if it does so, its balance is not enforced anymore. This is what happens on the following scheme:

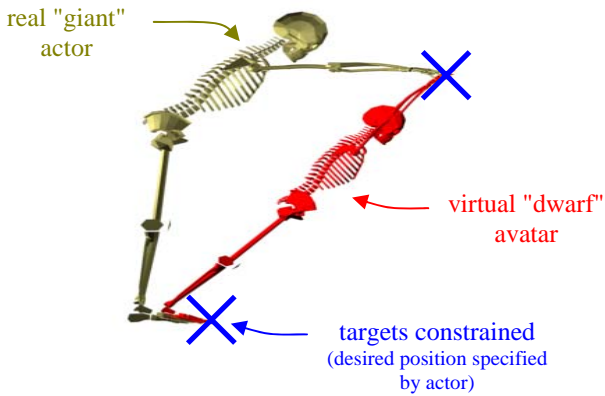

**Fig. 2 When retargeted, the giant's movement gives an unbalanced movement on the dwarf**

To account for such situations, Laszlo, van de Panne, and Fiume introduced Limit Cycle Control [5] which makes it able to compensate for small disturbances of a cycling movement (seen in a reduced state space). In the given example, the compensation for regulation variables called *up-vector* is made through control variables that are chosen to be the hip pitch and roll angle. This approach is interesting for cycling movements such as walking, but we want to be able to perform whatever movement. Hodgins and Wooten [6], proposed to control ankles and hips angles to enforce static equilibrium, but their approach is specific to one kind of movements (vaulting in their example). Faloutsos, van de Panne, and Terzopoulos [7] detect disturbances thanks to the famous notion of *support polygon*, and try to balance their virtual manikin thanks to the only ankle's stiffness correction. Although their approach tackles particularly well with situations where balance is lost, their control is rather restrictive.

When an actual human's balance is disturbed, the natural reaction is not to adjust a single or a couple of joints, the whole body is involved in the balance recovering.

Liu and Popovic [8], proposed to blend a distance to balance in the objective of a space-time optimization (ensuring an answer to disturbance distributed on all joints), enabling the generation of nice movements from sketches. Unfortunately, it suffers from the space-time optimization problem highlighted above. Fang and Pollard brought a physical filter [9], which does not allow real time performances, though it is much faster than previous attempts.

The biped robots community encounters the same kind of balance problem as we do. Besides the solutions already shown they also use the well known notion of Zero Moment Point (ZMP) [10], which allows to study dynamic equilibrium, but only when all contacts with environment occur on a plane (e.g. walking, running…). Harada et al. [11] extend the notion of ZMP to situations where contacts are not located on the same plane anymore, and ensure balance thanks to a method they propose, based on linear complementarity.

Our virtual humans control scheme is to support balance control in an interactive manner.

**Contact in virtual environment:**

The interest of virtual reality is to offer the possibility to evolve in virtual worlds; that is worlds that have no real counterpart. Knowing this, one can imagine the problems arising.

Imagine a virtual human is facing a virtual wall. In a usual motion capture session, the actor is not constrained in such a way. That is the real world actor can break the virtual constraint (that is the real actor can reach areas which virtual counterparts are occupied by the environment), whereas its avatar must not!

Haptic devices [12] provide an interesting approach, enabling to apply forces into the actor so as to prevent its avatar from penetrating the virtual environment. The framework being developed is fully compatible with such approaches. Nevertheless this method requires a heavy infrastructure.

Another way to deal with such kinds of problem is to implement a contact solver at the simulated world level. That is a controller that enforces environmental constraints in the simulated world, whatever the movements of the real world actor.

Zordan and Hodgins introduced "hitting and reacting" manikins [13], with a technique which principle is to modify control gains during the simulation. This method mainly aimed at games, is known to lack of stability, moreover, it is computationally heavy, and thus the solution is not real time.

Schmidl and Lin [14] implemented an approach based on both inverse kinematics - to solve for the manikin's reaction to contact -, and impulse-based physics for the environment. There hybrid approach lose the physical nature of the simulation.

The framework we propose allows to manage forces in real time simulations, under the natural physical laws. That is interaction with the environment will be natural.

The purpose of the present paper is the extension of an existing architecture. In [1], we proposed a passive control architecture that brought natural interactivity with environment through forces, retargeting, and that made it able to "help" the actor perform its movement[1] thanks to virtual guides, which is a semi-automatic command mode.

---

[1] **Help the actor perform its movement :** due to the lack of haptic sensations, the actor may have trouble to achieve given movements.

This control satisfies our industrial need for virtual prototyping, except when the actor, and its virtual counterpart are completely different. In this case we encounter the balance problem detailed above. Thus we aim at validating the possibility to balance virtual humans in our control architecture. This is the main feature of the present paper.

We will quickly describe the control architecture set-up, then we will introduce our balance control (mainly aimed at validating balancing), and finally show the results obtained.

## 2. Proposed architecture

The proposed architecture we developed can be seen on the following scheme.

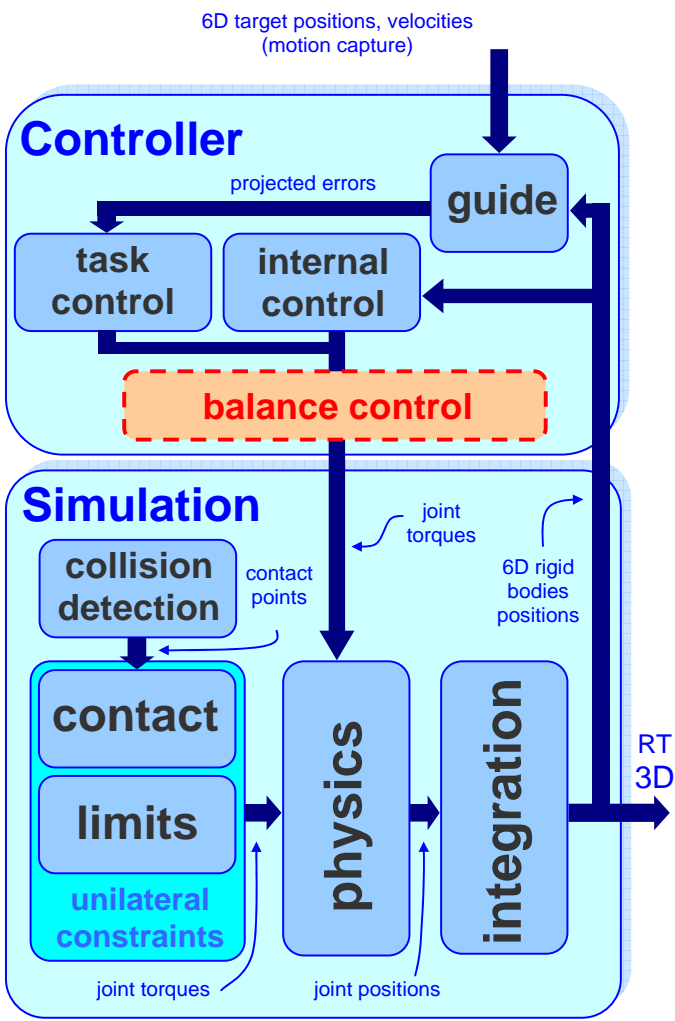

**Fig. 3 Detailed architecture**

---

Imagine a worker drilling a hole, in the real world, he is guided by the haptic sensation of the drill in the material (let's say wood). When no haptic feedback is available, we can use virtual guides to help him.

This control scheme distinguishes two main blocks: on the one hand we find the simulation, which emulates the real world's physical laws and the manikin, and on the other hand the controller drives the virtual human to the desired goal, expressed in terms of task space targets positions, and other constraints (such as balance, guides…). Now we rapidly detail the internal behavior of the control scheme.

Desired targets positions are received from a motion capture device. The error is projected in a passive way, thanks to mechanical analogies as explained in [1], hence creating virtual guides. Then the projected error goes through a task space corrector, which will generate a compensation for the error. As virtual humans are highly redundant systems, we can add to the task space control an internal control (which must not interfere with task space control).

Now comes the management of unilateral constraints. As we will see in section 3., balance control can be seen as a unilateral constraint enforcing the balance of the virtual human being constrained. Joint limits and contact response are managed in the *Simulation block*, because they are not specific to character animation, they respectively enforce joint limits (of course), and the non penetration with environment, but also the interaction with environment: that is virtual humans can apply forces onto the virtual environment, hitting, pushing, and pulling as a real human would do on a real environment...

After that, we us GVM, and LMD++, two packages developed by CEA\LIST, to perform physical simulation [15].

Now we describe the main innovation of the present paper: the introduction of the balance controller.

## 3. Balance control

Our main purpose was to check, thanks to a simple controller, if our architecture could handle balance control. That is why we chose to build the balance controller on the well known concept of support polygon.

This notion states that *walking systems remain statically balanced so long as the vertical projection of their center of mass stays inside the convex hull of contact points* [16].

Balance can be seen as a unilateral constraint rather straightforwardly. The unilateral constraint solution is given by the resolution of a Linear Complementarity Problem (LCP). Its superiority with respect to regulation methods is detailed in [17].

A general LCP can be expressed as follows:

*Ensure unilateral constraints* $\omega \geq 0$, *and* $z \geq 0$, *knowing complementarity* $\omega^T z = 0$, *and the relation between* $\omega$, *and* $z$: $\omega = Mz + q$.

Usually expressed in a shorter shape: $0 \leq \omega \perp z \geq 0$, with $\omega = Mz + q$. $\omega$ can be seen as a control variable, and

$z$ as the distance to constraint, further explanations can be found in [18].

We now express our balance problem as a LCP. We will illustrate our approach on a virtual manikin standing straight, with both feet on the ground. We know the projected center of mass must lie inside the support polygon. As seen on Fig. 4, we will approximate the support polygon of both feet by an ellipse, this is done without loss of generality, because the LCP could be expressed with a polygon as well.

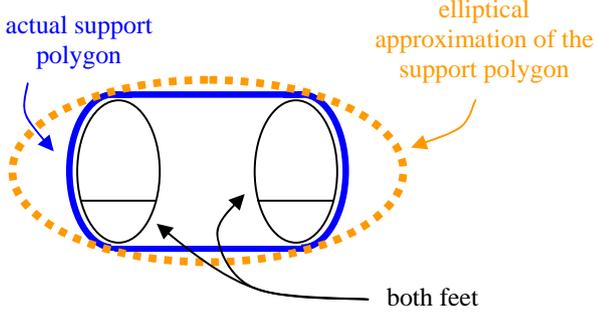

**Fig. 4 Elliptical approximation of the actual support polygon**

Our support polygon approximation makes it able for us to regard the configuration of a virtual human as balanced when the vertical projection of the center of mass lies inside the elliptical limit, as seen on Fig. 5:

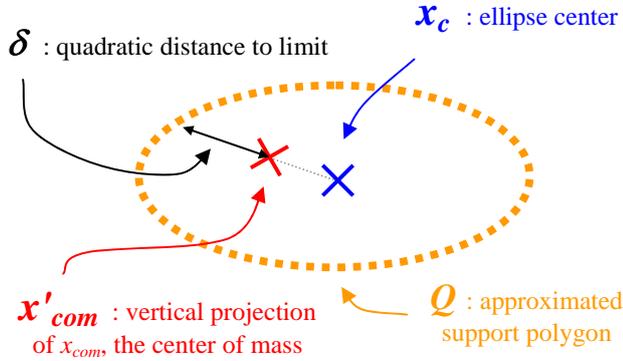

**Fig. 5 Schematic diagram of our balance control: we aim at keeping $x_{com}$ inside of $Q$.**

As for our problem, the LCP, is expressed as follows:

We want $\delta$, the quadratic distance to limit, to remain positive or zero, if this constraint breaks, the LCP solver, will modify $\Gamma_{LCP}$ (the joint torques due to the unilateral constraint enforcement) to enforce the constraint on $\delta$, this modification being done according to the relation between $\delta$ and $\Gamma_{LCP}$.

In [15], it is shown that this relation between $\delta$ (the variable being constrained), and $\Gamma_{LCP}$ (the control variable), can be expressed by the Jacobian matrix of $\delta$, with respect to $q$ (the joint parameters of our virtual human), when the equation of evolution of the system (the dynamics equation) is known. That is the LCP solver's input will be the Jacobian matrix of $\delta$. Thus we now express $\delta$, and its Jacobian matrix.

As $Q$, is an ellipse, $\delta$ can be written as:
$$\delta = d^2 - \|P(x_{com} - x_c)\|_Q^2, \quad (1)$$

with $d$, the maximum distance, $P$, the vertical projection, and $Q$, the metric corresponding to the ellipse $Q$.

$J$, the Jacobian matrix of $\delta$, can be expressed by:
$$J = \frac{\partial \delta}{\partial q}, \quad (2)$$

thanks to eq. (1), $\partial\delta$ becomes:
$$\partial\delta = -\partial\|P(x_{com} - x_c)\|_Q^2, \quad (3)$$

considering the virtual human is not changing its support polygon (that is double feet support, is made independent of single foot support, during a walk), we have:
$$\begin{aligned}\partial\delta &= -(x_{com}-x_c)^T P^T (Q+Q^T)\partial(P(x_{com}-x_c))\\ &= -(x_{com}-x_c)^T P^T (Q+Q^T)P\partial x_{com}\\ &= -(x_{com}-x_c)^T P^T (Q+Q^T)PJ_{com}\partial q,\end{aligned} \quad (4)$$

with $J_{com}$ being the Jacobian matrix of the center of mass. We know have to express $J_{com}$.

The position of the center of mass $p_{com(0)}$ of an articulated system, expressed in the base frame $0$, is given by:
$$p_{com(0)} = \frac{\sum m_i p_{com\_i(0)}}{\sum m_i}, \quad (5)$$

with $p_{com\_i(0)}$ being the position of the center of mass of the $i^{th}$ solid, thus the velocity $v_{com/0(0)}$ of the full system's center of mass seen from base frame $0$, and expressed in the same frame is:
$$v_{com/0(0)} = \frac{\partial p_{com(0)}}{\partial t} = \frac{\sum m_i \frac{\partial p_{com\_i(0)}}{\partial t}}{\sum m_i}$$
$$= \frac{\sum m_i v_{com\_i/0(0)}}{\sum m_i}$$

so $J^r_{com/0(0)}$ (written $J_{com}$ in eq. (4)), the Jacobian matrix of $v_{com/0(0)}$ is given by:
$$J^r_{com/0(0)} = \frac{\sum m_i J^r_{com\_i/0(0)}}{\sum m_i}, \quad (6)$$

$J^r_{com/0(0)}$ is a reduced Jacobian matrix, because a center of mass position has 3 translational components, whereas a full solid position is 6D (3 rotations more), the Jacobian matrix associated to a general solid position is written $J_{A\in i/j(k)}$

(one must understand Jacobien of the speed of point *A* fixed in frame *i*, in its movement with respect to frame *j*, expressed in base *k*). The relation between $J_{A\in i/j(k)}$, and $J^r_{A\in i/j(k)}$ is given by:

$$J^r_{A\in i/j(k)} = \begin{pmatrix} 0_{3*3} & I_3 \end{pmatrix} J_{A\in i/j(k)} = SJ_{A\in i/j(k)} \quad (7)$$

Knowing this, we can now express $J^r_{com/0(0)}$, thanks to the Jacobien matrices of the center of mass of each solid of the articulated system $J_{com\_i/0(0)}$. Eq. (6) gives us:

$$J^r_{com/0(0)} = \frac{\sum m_i J^r_{com\_i/0(0)}}{\sum m_i}$$

$$J_{com} = J^r_{com/0(0)} = \frac{\sum m_i S J_{com\_i/0(0)}}{\sum m_i} \quad (8)$$

Introducing $\delta$, and $J_{com}$, into our LCP solver (which choice is out of scope) gives the interesting results seen in next section.

## 4. Results

The balance controller we designed is aimed at enforcing static equations, in a framework allowing real time animation, interaction with environment, virtual guides... Its great behavior is illustrated bellow, first we show figures of the system's behavior to collision:

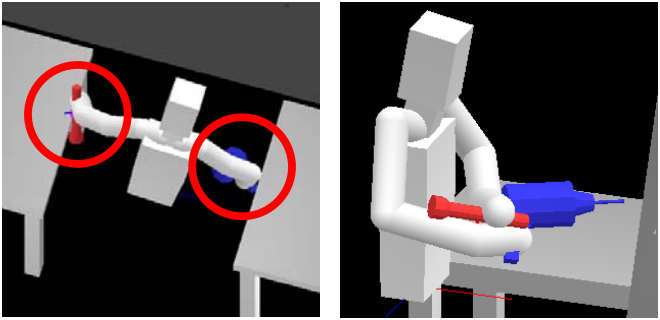

**Fig. 6 Double and self collision**

On the curve Fig. 7, we can see the height of the table, which must not be penetrated (dashed orange), and the height of the virtual human's hand (green), while reaching, and leaning on the table: the hand never penetrates the table.

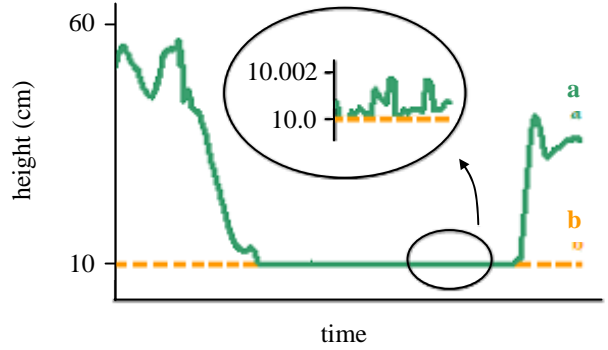

**Fig. 7 Hand (a), and obstacle's (b) height: no penetration.**

We now test the virtual guides approach we have implemented. The experiment consists in drilling a hole in a wall thanks to a drill, while lighting the future hole's location thanks to a hand light. The drill can only move along a fixed axis with a fixed orientation. This means that the controller leaves only one degree of freedom to the operator. The direction of the spotlight is also driven automatically (leaving the three degrees of freedom of the light's position to the operator). Fig. 8 depicts the ideal axis in green (a) and actual axis are in red (b).

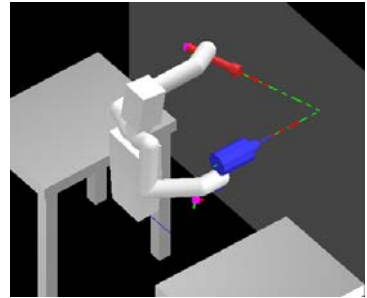

**Fig. 8 Worker drilling a hole, guided by virtual mechanisms**

In order to see the efficiency of our method, we drew the angle between the ideal axis, and the actual axis of the drill, as seen on Fig. 9; in the case where the operator is completely free (green), and in case where the guide is on (dashed orange).

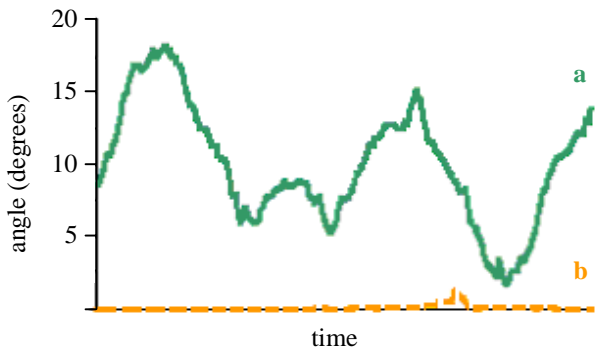

**Fig. 9 Angle between ideal and actual axis of the drill, (a) without guide, and (b) with guide.**

Now we show the balance controller's action:

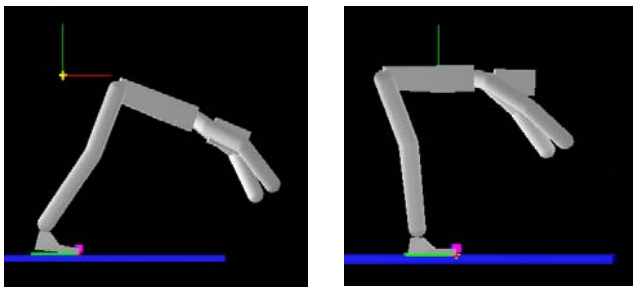

**Fig. 10 Unbalanced dwarf being controlled (left image), and the same dwarf being balance controlled, while the giant actor performs the same movement (right image).**

We can see that the configuration proposed by the unbalanced controller is unfeasible, whereas with be balance controller on, the system behaves well: a real human could adopt this posture without falling.

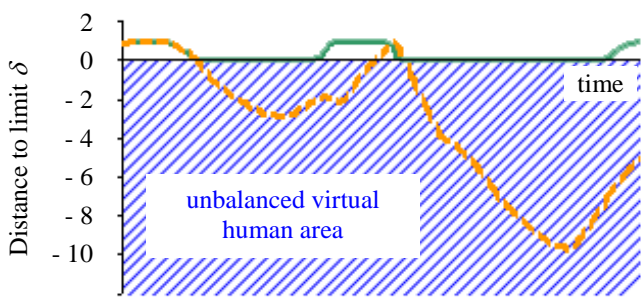

**Fig. 11 Distance to limit in the unbalanced case (dashed orange), and in the balanced case (green). The distance to limit is expressed in multiples of *d*.**

Fig. 11 shows the constraint is sharply enforced in case of balance control (green line), which is not the case when no control is done on balance (dashed orange line).

We should notice that the model we chose as for balance (observing the projection of the center of mass), can only enforce static balance of walking avatars, that is the balance model is correct so long as the manikin does not interact with environment in another way than with the feet, on a plane surface.

## 5. Conclusion

In this paper, we proposed a control architecture enabling to animate a virtual human in real time in an immersive way, thanks to a motion capture device. This architecture makes it able to interact with the environment through forces. Moreover we added the possibility to enforce static balance while retargeting the motion to an avatar which morphology is different than the actor's one.

Our main purpose was to validate the possibility to balance virtual humans on our control architecture, thanks to a simple balance model.

Hence the natural continuation will be to extend the balance controller possibilities to handle multi contact with friction.

**Bibliography**

[1] A. Rennuit, A. Micaelli, X. Merlhiot, C. Andriot, F. Guillaume, N. Chevassus, D. Chablat, P. Chedmail, "Passive Control Architecture for Virtual Humans", submitted to *IEEE/RSJ International Conference on Intelligent Robots and Systems IROS'2005*, august 2005, Edmonton, Canada.
[2] M. Gleicher, "Retargetting Motion to New Characters", *Proceedings of SIGGRAPH 98*, in Computer Graphics Annual Conferance Series. 1998.
[3] Z. Popovic, A. Witkin, "Physically Based Motion Transformation", *Siggraph 1999*, Computer Graphics Proceedings.
[4] P. Baerlocher, R. Boulic, "An Inverse Kinematic Architecture Enforcing an Arbitrary Number of Strict Priority Levels", *The Visual Computer*, Springer Verlag, 20(6), 2004.
[5] J. Laszlo, M. van de Panne, E. Fiume, "Limit Cycle Control, and its Application to the Animation of Balancing and Walking", *Proceedings of ACM SIGGRAPH*, New Orleans, Louisiana, August 1996.
[6] J.K. Hodgins, W.L. Wooten, "Animating Human Athletes", in *Robotics Research: The Eighth International Symposium*, 1998. Y. Shirai and S. Hirose (eds). Springer-Verlag: Berlin, 356-367.
[7] P. Faloutsos, M. van de Panne, D. Terzopoulos, "The Virtual Stuntman: Dynamic Characters with a Repertoire of Autonomous Motor Skills", *Computers and Graphics*, Volume 25, Issue 6, December, 2001, pp. 933-953.
[8] C.K. Liu, Z. Popovic, "Synthesis of Complex Dynamic Character Motion from Simple Animations", *Proceedings of ACM SIGGRAPH, July 2002*, San Antonio USA.

[9] A.C. Fang, N.S. Pollard, "Efficient Synthesis of Physically Valid Human Motion", *Proceedings of ACM SIGGRAPH, July 2003*, San Diego, USA.

[10] M. Vukobratovic, B. Borovac, "Zero Moment Point – Thirty Five Years of Its Life", *International Journal of Humanoid Robotics*, Vol. 1, No. 1 (2004) 157-173.

[11] K. Harada, H. Hirukawa, F Kanehiro, K. Fujiwara, K. Kaneko, S. Kajita, M. Nakamura, "Dynamical Balance of a Humanoid Robot Grasping an Environment", *Proceedings of 2004 IEEE/RSJ International Conference on Intelligent Robots and Systems IROS'2004*, October 2004, Sendai, Japan.

[12] P. Garrec, J.P. Martins, J.P. Friconneau, "A new technology for portable exoskeletons", submitted to AMSE : *International Association for Modelling and Simulation*.

[13] V. B. Zordan and J. K. Hodgins. Motion capture-driven simulations that hit and react. In *Proceedings of the 2002 ACM SIGGRAPH/Eurographics symposium on Computer animation*, pages 89–96. ACM Press, 2002.

[14] H. Schmidl and M. C. Lin, "Geometry-driven Interaction between Avatars and Virtual Environments", *Computer Animation and Virtual Worlds Journal*, July 2004.

[15] X. Merlhiot, "Dossier d'analyse des algorithmes embarqués", *CEA Internal Technical Report*, published in the framework of the LHIR RNTL project, January 2005.

[16] P.B. Wieber, "On the Stability of Walking Systems", *Proceedings of the International Workshop on Humanoid and Human Friendly Robotics*, 2002.

[17] C. Duriez, "Contact frottant entre objets déformables dans des simulations temps-réel avec retour haptique", *PhD thesis*, Université d'Evry, December 2004.

[18] K.G. Murty, F.T. Yu, "Linear Complementarity, Linear and Nonlinear Programming", *Heldermann Verlag*, Berlin, 1988.